\documentclass[10pt,twocolumn,letterpaper]{article}

\usepackage{iccv}
\usepackage{times}
\usepackage{epsfig}
\usepackage{graphicx}
\usepackage{amsmath}
\usepackage{amssymb}
\usepackage[utf8]{inputenc}

\usepackage{graphics} 
\usepackage{epsfig} 
\usepackage{mathptmx} 
\usepackage{times} 
\usepackage{amsmath} 
\usepackage{amssymb}  
\usepackage{siunitx}
\usepackage{tabularx}
\usepackage{authblk} 

\usepackage[pagebackref=true,breaklinks=true,letterpaper=true,colorlinks,bookmarks=false]{hyperref}

\iccvfinalcopy

\def\iccvPaperID{32} 

\ificcvfinal\fi

\newcommand{\cmmnt}[1]{\ignorespaces}

\DeclareMathOperator*{\argmax}{arg\,max}

\title{Correcting Decalibration of Stereo Cameras in Self-Driving Vehicles}

\author[1]{Jon Muhovi\v{c}\thanks{jon.muhovic@fe.uni-lj.si}}
\author[1]{Janez Per\v{s}\thanks{janez.pers@fe.uni-lj.si}}
\affil[1]{Faculty of Electrical Engineering, University of Ljubljana}

\begin{document}

\maketitle
\ificcvfinal\thispagestyle{empty}\fi

\begin{abstract}

We address the problem of optical decalibration in mobile stereo camera setups, especially in context of autonomous vehicles. In real world conditions, an optical system is subject to various sources of anticipated and unanticipated mechanical stress (vibration, rough handling, collisions). Mechanical stress changes the geometry between the cameras that make up the stereo pair, and as a consequence, the pre-calculated epipolar geometry is no longer valid. Our method is based on optimization of camera geometry parameters and plugs directly into the output of the stereo matching algorithm. Therefore, it is able to recover calibration parameters on image pairs obtained from a decalibrated stereo system with minimal use of additional computing resources. The number of successfully recovered depth pixels is used as an objective function, which we aim to maximize. Our simulation confirms that the method can run constantly in parallel to stereo estimation and thus help keep the system calibrated in real time. Results confirm that the method is able to recalibrate all the parameters except for the baseline distance, which scales the absolute depth readings. However, that scaling factor could be uniquely determined using any kind of absolute range finding methods (\eg a single beam time-of-flight sensor).

\end{abstract}

\section{Introduction}
All stereo camera systems require calibration before they are ready to obtain 3D measurement from the visual data. In the context of stereo vision, calibration is a process, where a target of precisely known dimensions is observed by both cameras, and a number of parameters that define transformation from world coordinates $(X,Y,Z)$ to image coordinates in left $(u_l,v_l)$ and right camera $(u_r,v_r)$ is calculated. During the calibration process, the intrinsic parameters of each camera and the physical relationship between the cameras (\ie the extrinsic parameters) have to be calculated.

\subsection{Calibration of the vehicle's stereo rig in practice}

\begin{figure}[ht]
    \centering
    \includegraphics[width=\columnwidth]{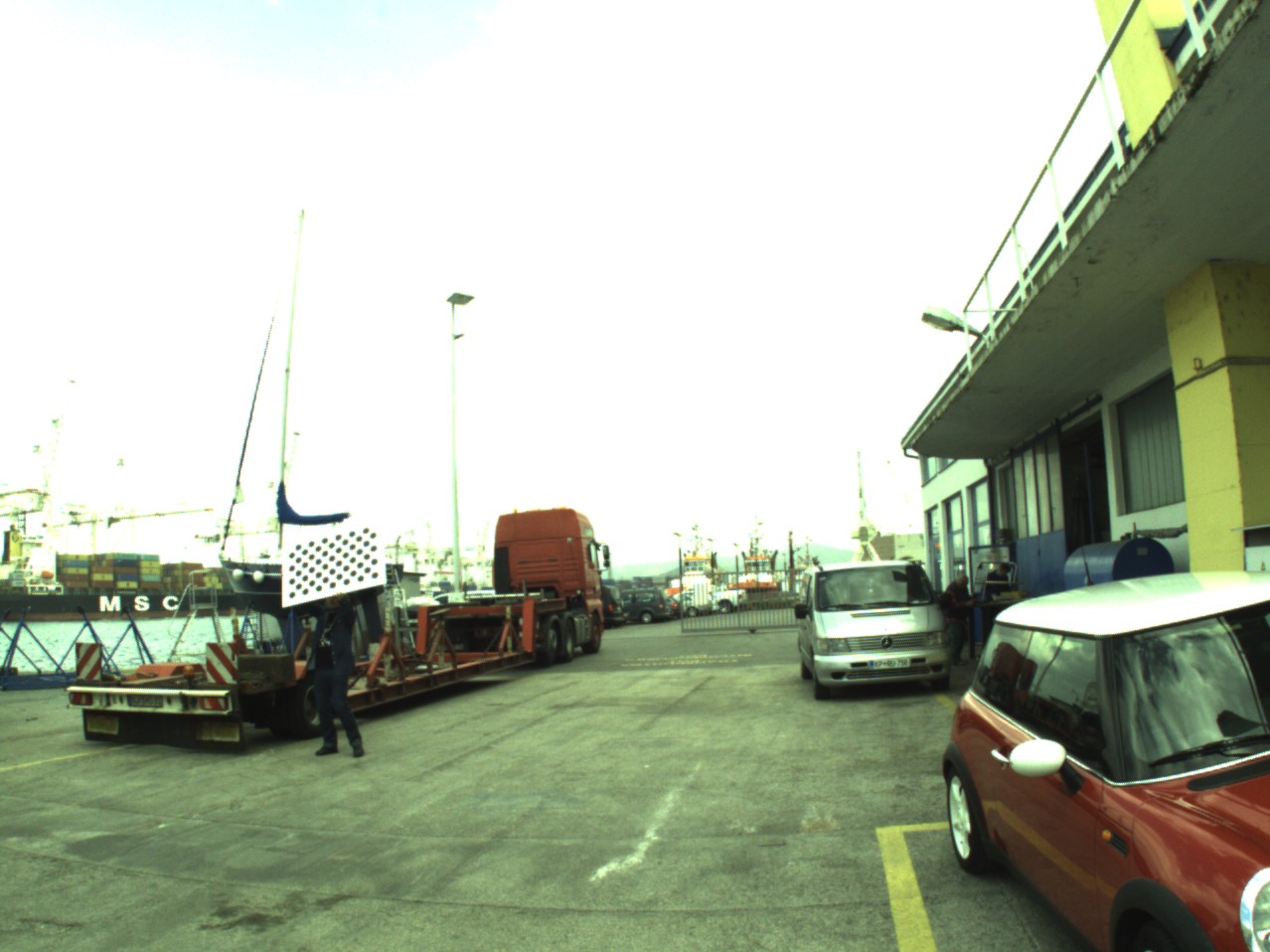}
    \caption{Calibration of a stereo system that is mounted on an unmanned surface vehicle (USV) is a completely different endeavour than calibrating the stereo rig in the laboratory.  Note that ''in the wild'' it is practically impossible to cover the whole image area with the calibration target, unless one of the operators is able to climb while holding the target.}
    \label{fig:outdoor}
\end{figure}

 In theory, calibration can be performed in a controlled environment with good lighting and by using a known pattern such as a checkerboard or asymmetric circle pattern (the latter is shown in~Figure~\ref{fig:outdoor}). But in mobile robotics and self-driving vehicles, this may be highly impractical, as shown in~Figure~\ref{fig:outdoor}. In this case, before calibration, the unmanned surface vehicle (USV) has been taken out of the water, and placed to face the area where the target can be moved to cover the whole range of the stereo system.

The differences in calibration between theory and practice arise from the following real-world constraints:
\begin{itemize}
\item The real-world stereo system that benefits self-driving vehicles is usually scaled up significantly to provide the useful range. That means a much wider baseline, and with that, higher sensitivity to mechanical stress.
\item Sensors in self-driving vehicles are subjects to size, weight and price constraints. It is not possible to use heavy, sturdy construction in all cases. The setup on our USV is shown in~Figure~\ref{fig:system} and has been mandated by constraints arising from the fact that our USV is essentially a boat, and centre of gravity should be low as possible.
\end{itemize}

Note that these constraints affect mainly the stability of extrinsic parameters. Intrinsic parameters are ''hidden'' inside each camera-lens system and can be proofed against accidental decalibration much more effectively (e.g. sealing with glue). Therefore, this paper deals with decalibration of external parameters only.

\subsection{Sources and consequences of stereo decalibration}

If the stereo system is one of the primary navigation tools for an autonomous vehicle, the undetected decalibration is extremely dangerous, as it may cause the vehicle to go "blind". The first ("factory") calibration cannot be avoided, but the movement, temperature changes and handling, even in storage, can cause decalibration at a later time. In vehicles that move over rough terrain or have other sources of vibration (\eg heavy-duty internal combustion engines in trucks, tractors and boats), the decalibration may be a regular occurrence, especially given the constraints placed upon the design of a stereo rig.

\begin{figure}[ht]
    \centering
    \includegraphics[width=\columnwidth]{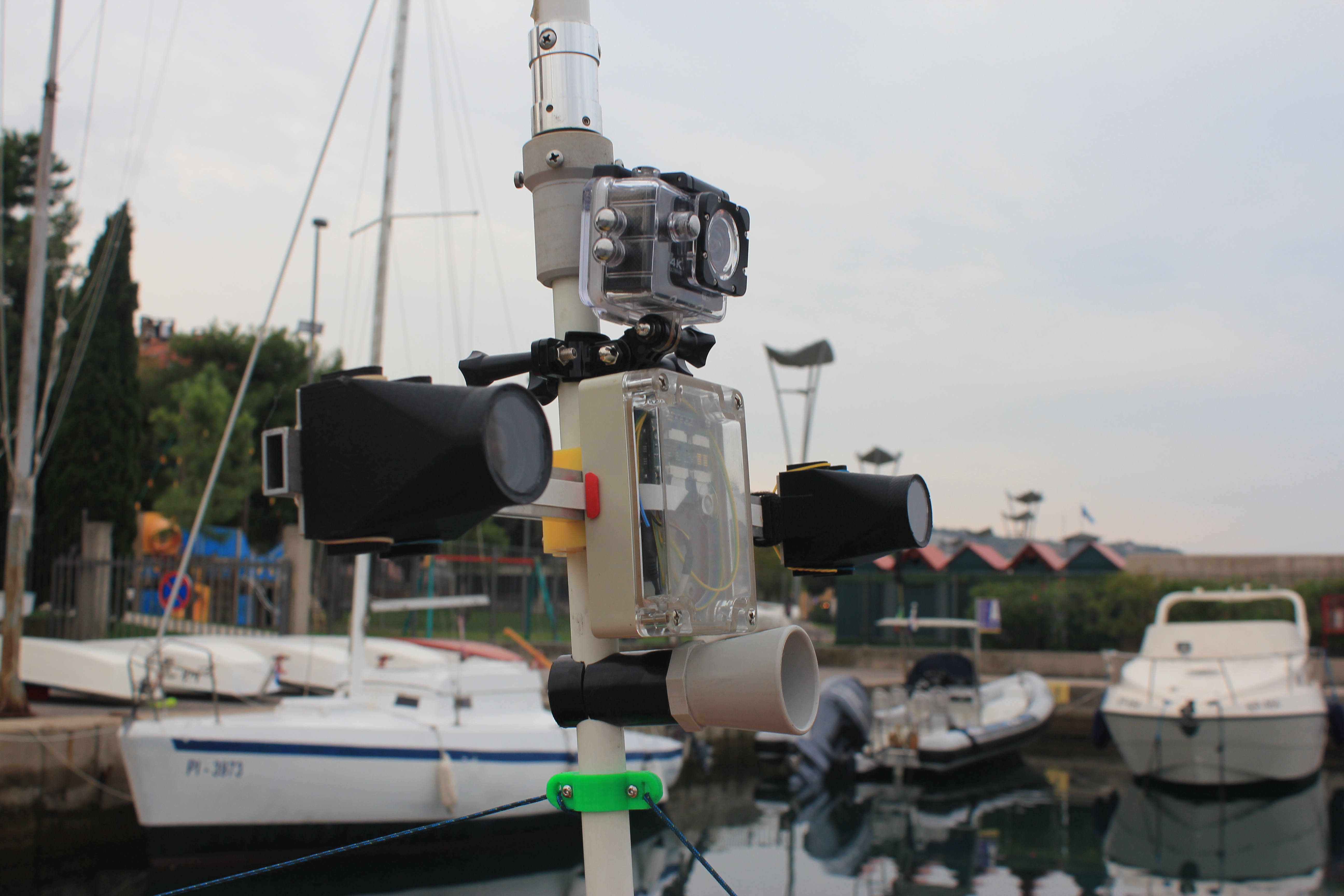}
    \caption{Our stereo system mounted on the USV.}
    \label{fig:system}
\end{figure}

The authors have a longstanding involvement with the development of an obstacle avoidance system for the unmanned surface vehicle (USV, shown in Figure~\ref{fig:system}), and have witnessed several occurrences of decalibration, which was revealed only after inspection of captured data. \emph{The symptom of decalibration} is chiefly heavily reduced number of stereo matches, which puts the vehicle in the extremely dangerous condition -- obstacles may not be detected at all. Manual calibration before each USV mission, as shown in~Figure~\ref{fig:outdoor} is time consuming and might not be possible due to lighting conditions, limited battery life, weather, or time constraints. In vehicles operated by users without engineering knowledge, the full calibration procedure including target movement is outright impossible. 


To address these problems, a method to detect decalibration and automatically correct a decalibrated system is needed. Such an automatic method could significantly simplify the deployment of mobile robots and increase safety of self-driving vehicles, and driver-assistance systems. Even a poor, but working calibration would be a better solution than a system that is blind and non-functional.

\section{Related work}
Geometry of stereo calibration has been studied in detail decades ago, therefore many relevant publications date way back. Additional push into research on stereo happened several years ago with realistic prospect of driver assistance and obstacle detection systems, with ultimate goal of developing fully self-driven vehicles.

The base for stereo vision methods was established as early as 1981 by Longuet-Higgins~\cite{longuet1981computer}, which describes the widely known 8-point algorithm for reconstructing a 3D scene from point matches in images. Later works, such as Hartley's \cite{hartley1995defence} and Nistér's~\cite{nister2004efficient} expand upon Longuet-Higgins' work and present improvements on the solutions of the pose reconstruction problem. 

The methods for automatic camera and stereo system calibration are also well documented and researched. Research regarding the estimation of the intrinsic and extrinsic parameters of cameras was published in papers like \cite{zhang1996motion, zhang2000flexible}. These set up the standard approach of matching the detected features in camera images to the patterns that are known upfront and thus producing a system of linear equations that can be solved with a closed form solution. Non-linear lens distortion parameters can also be estimated in a similar way. The work of Takahashi \etal~\cite{takahashi1988self} deals with estimating the rotation parameters of a stereo system given the observed points which are calculated via boundary representations of input images. The papers by Dang \etal~\cite{dang2006self, dang2009continuous} explore the calibration of a stereo system with movable cameras by using bundle adjustment on tracked points in sequential images combined with epipolar constraints. The calibration is then continuously optimized using an Iterated Extended Kalman Filter (IEKF). The work by Collado \etal~\cite{collado2006self} deals with determining the height, roll and pitch of a stereo system used for driver assistance systems. Here, the road lanes detected in the images captured from a car mounted stereo system are used as a calibration pattern under assumptions on planarity and parallelism of the lanes on a straight road segment. 

Research into self-driving technology introduced concepts that have not been popular before, such as systems with heterogenous cameras. Those need adapted calibration targets and different methodology for efficient (practical) calibration, as for example~\cite{Rathnayaka2017}. The introduction of high definition cameras and related inexpensive hardware had a similar effect, which called for efficient calibration of high-definition stereo rigs~\cite{Chen2012}. The consequence of stereo imaging systems entering widespread use is indeed the search for methods that are easy to operate and still provide high-quality stereo calibration, one of such methods is presented in work by Sun \etal~\cite{Sun2018}. Interesting common point with our system is that they specifically target large calibration volumes, which cannot be efficiently calibrated by the conventional checkerboard pattern methods. The acknowledgement that fiducial marker-based (checkerboard) approaches to calibrate multiple cameras are impractical in the field of self-driving vehicles, is the reliance on natural features through the SLAM based approaches, as in~\cite{HANE2017}. Motivation for our approach is somewhat similar.

Finally, some existing methods use additional sensors to complement the camera system or even seek to calibrate sensors with respect to the camera. In~\cite{Moravec2018}, authors describe a system which calibrates the stereo and LIDAR without the use of calibration targets, relying only on the observed surroundings of the vehicle.

Our approach to performing adjustments to calibration is \emph{introspective}, which means that it does not need additional sensors -- it performs optimization on acquired image data only. It can be argued that it is simple and self-evident, but to best of our knowledge, the method has not been published before.

\section{Proposed method}

In this section we present a short overview of the principles of epipolar geometry and stereo vision. Even though our method is not complex, epipolar constraints serve as foundation for our method.

\subsection{Epipolar geometry}
If a point $X$ in 3D space is observed by a single camera, the coordinates of the point image $X_L$ on the image plane are known, but the point's distance to the image plane is not. The point could lie anywhere on the beam from the camera center to infinity. If another camera then observes the same point, the coordinates of the point's image ($X_R$) in the second camera are constrained by the physical relationship of both cameras. If the poses of both cameras are known, the position of a point's projection in one camera also gives us a line on which the point's image in the second camera must lie. This line is known as an epipolar line.
A special case of epipolar geometry is the case when cameras have the same orientation and are displaced only on one axis. This causes the epipolar lines for all points in both images to be parallel, which means that matching points lie on the same image row. Stereo matching algorithms exploit this property to reconstruct the 3D scene observed by the cameras. This is derived from the fact that points closer to the cameras will be displaced more than those far away from the cameras (this is known as disparity). Stereo systems use disparity to infer the depth of image points seen with both cameras. Of course perfect alignment of cameras is impossible to achieve in reality. Instead, minute displacements are corrected using image rectification that virtually aligns both image planes.

\emph{The key idea} of our method is the observed fact that violation of epipolar constraints as a consequence of decalibration inevitably \emph{causes stereo algoritems to detect fewer matches}. They are optimized to search for matches along epipolar lines only, and the greater the decalibration, the fewer matches will be detected!

\begin{figure}[htb]
    \centering
    \includegraphics[width=0.9\columnwidth]{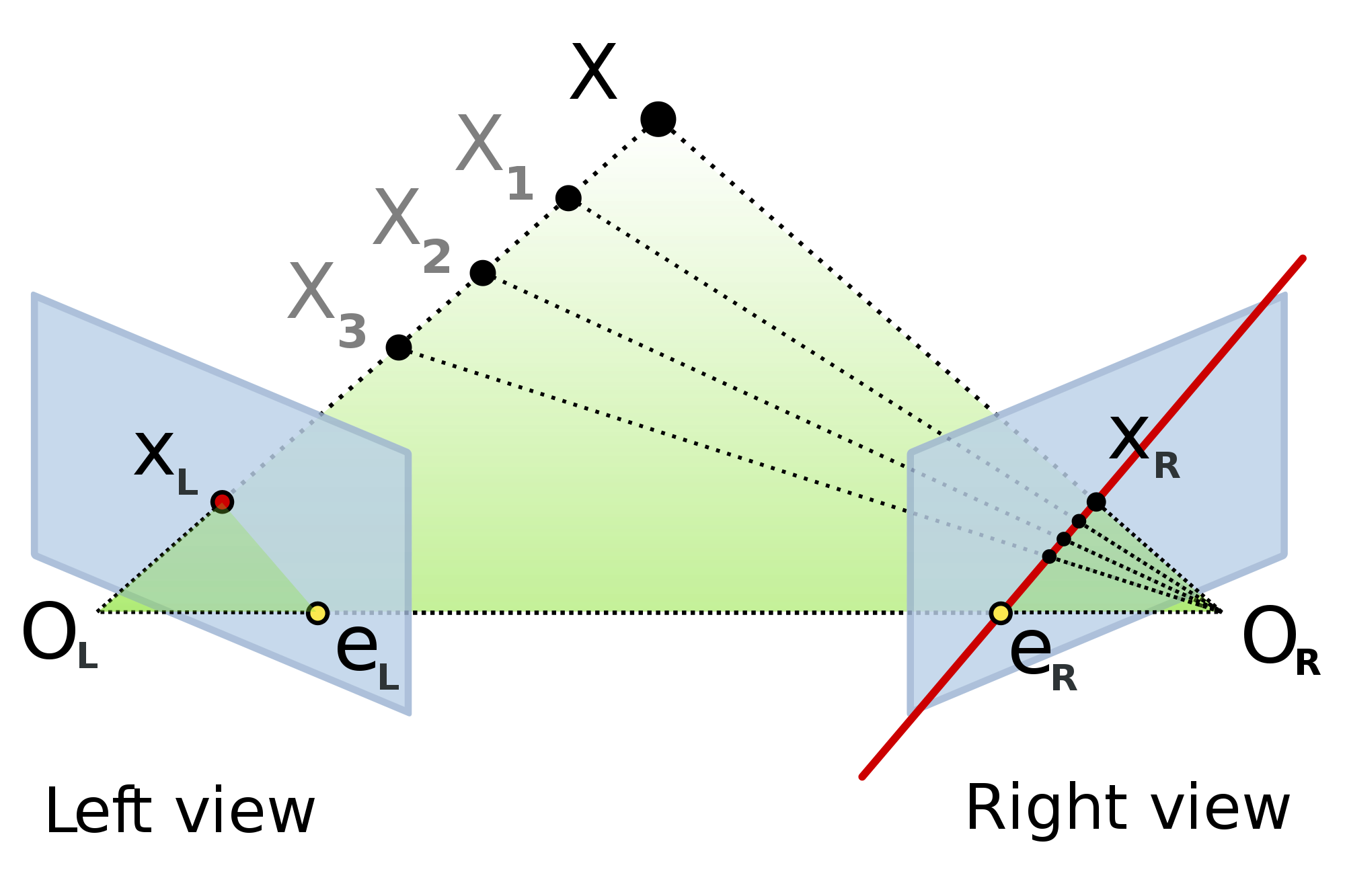}
    \caption[]{Visual representation of general epipolar constraints.\protect\footnotemark}
    \label{fig:epipolar}
\end{figure}

\footnotetext{Source of the image: Wikipedia. Reproduced under Attribution-ShareAlike 3.0 Unported (CC BY-SA 3.0) license. Image author: Arne Nordmann}

\subsection{Calibration and decalibration}

The usual assumption in stereo systems is that the cameras have the same orientation, and are displaced in exactly one dimension (the displacement between cameras is known as "baseline distance"). Usually, the displacement is on the $x$ axis, which allows stereo matching algorithms to exploit the properties of epipolar geometry.

To obtain the minute translations and rotations of the system that are caused by imperfect mounting and operational vibrations, calibration using calibration targets such as checkerboards is performed. In the paper, we will refer to this calibration as "manual", even though it is fairly automatic, but still requires some level of human involvement. The most tedious task in manual calibration is changing the location of the target so that it covers most of the area where we want to perform 3D reconstruction. In self-driving vehicles that may be huge (tens of meters in front of the vehicle, and 10-30 meters in width).

Due to movement, temperature changes and vibrations during operation, the physical relationship of the cameras in a stereo system can change. In that case, the manual calibration no longer adequately describes the system properties. This causes the epipolar lines in the images to no longer be parallel, which violates the assumption of stereo matching algorithms. Such a system starts to perform poorly and recalibration is needed.

If a manual calibration is not possible, a method of automatic recalibration should be devised. A direct calibration using feature points and the essential matrix is theoretically possible, but due to a various environments the USV operates in, might not be viable or reliable. The images captured by our system contain the water surface, which (when small waves are present, generating enough texture for stereo matching) can be used for stereo methods, but proves too noisy for general purpose feature detection and matching.

\subsection{Our approach}
\label{approach}

We propose to use the number of pixels recovered by a stereo matching algorithm as an indirect measurement of the calibration quality. The premise is fairly simple: the stereo matching algorithm assumes image alignment. If that is violated, the performance of the algorithm suffers, under the assumption that parameters of the stereo method remain the same (which, in realistic scenarios, is always true). Optimization of the extrinsic parameters of the system then strives to repair the alignment and thus allow the stereo matching to perform better. To the best of our knowledge, such an approach has not yet been done, despite its relatively simple premise.

The image rectification in a stereo system is based on the intrinsic and extrinsic parameters of the cameras. We assume that both cameras' intrinsic parameters were previously estimated by a manual calibration and are fixed (though this might not hold for very stressful environments). The errors that occur due to decalibration are then fully included in the extrinsic parameters. The procedure we describe here might be used to calibrate some of the intrinsic parameters that are prone to decalibration as well (most notably the center of the image plane, and perhaps focal length), but these are beyond the scope of this paper and have not been included in our experiments.

Stereo matching methods require two rectified images as input. A one-dimensional search along the epipolar lines, which are perfectly horizontal after rectification, is then performed for each pixel in the left image to find the location of the visually most similar pixel in the right image. The resulting image is called the disparity im

The set of extrinsic parameters that define a stereo calibration can be written as
\begin{equation}
    \Upsilon = [\vartheta, \varphi, \psi, x, y, z ],
\end{equation}
where $\vartheta$, $\varphi$, $\psi$ are the pitch, yaw and roll angles respectively. Parameters $x$, $y$ and $z$ are elements of the translation vector $T$.
The cost function $f: \mathbb{R}^6 \rightarrow \mathbb{N}$ can then be written as
\begin{equation}
    f(I_l,I_r,\Upsilon),
    \label{eq:f}
\end{equation}
where $I_l$ and $I_r$ are the left and the right image of the stereo system. As the left camera center is assumed to lie in the origin of the coordinate system, the parameters of the 3D rigid transformation define the position of the right camera in relation to the left one.

This sets up the optimization problem as finding the point in 6D space that maximizes the number of valid disparities based on a stereo image pair. Formally, we are searching for
\begin{equation}
    \argmax_{\Upsilon \in \mathbb{R}^6} f(I_l,I_r,\Upsilon).
\end{equation}

While generally 3D pose has 6 degrees of freedom, we fix the offset along $x$ axis to our system baseline (which is known) because it does not affect the stereo quality, but only the scale, which depends on the stereo system baseline. If we were to use a depth sensor to constrain the baseline distance, the parameter $x$ could also be optimized.

It must be noted that our approach does not necessarily converge to the same extrinsic position as the manual calibration. Since the manual calibration is not perfect, a better solution (both in terms of our metric and visually) may be found. While it is true that solutions that deviate from the ground truth may be incorrect, we did not notice any obviously degenerate solutions when observing the disparity images produced by optimized calibrations.

\subsection{Optimization methods}
As the function $f$ cannot be optimized analytically, numerical optimization methods were used.
Variants of the gradient descent method were tried at first, with the required gradient calculated as the vector of numerical derivative approximations. Additionally, a secant method that attempts to approximate the Hessian matrix was also used. The methods performed fairly well when the decalibration was limited (\ie when sufficient initial information was available). When the initial distance to a good calibration was larger (when we applied synthetic noise) gradient methods did not consistently perform well. If we fine-tuned optimization parameters, good results could be obtained, but a more robust method was needed.

A pattern search method called \textit{compass search} \cite{kolda2003optimization} that does not require gradient approximations was then used. Pattern search algorithms are used for optimizing functions that are not continuous or differentiable. The idea of the compass search is as follows: in each iteration, each of the parameters is increased and decreased by the same value, then the cost function is evaluated in those points. This produces $2n$ points where $n$ is the dimensionality of the problem. The estimate then moves into the point with the highest function value (if maximizing). If none of the calculated values is better than the current position, the step size is halved and the algorithm moves to the next iteration.

In Section \ref{sec:exp} we show that compass search works well for our problem, converges relatively fast and does not require fine tuning of parameters as can be the case with gradient descent methods.

\section{Experiments}
\label{sec:exp}
In the period of working on the USV that is equipped with stereo camera system (spanning 7 years), manual stereo calibration using calibration targets was performed 11 times. Usually, the moment for manual calibration was chosen after the performance of the stereo method degraded noticeably. While analyzing the extrinsic parameters of the calibrations we observed that the differences between them pre- and post-calibration could reach as high as 2.5$^{\circ}$ in rotation and up to \SI{8}{\milli\meter} in translation. This shows that decalibration both during the vehicle operation and due to storage can be significant. We also have image data from 39 USV sessions available and are thus able to retroactively check the calibration quality. For the purposes of the experiments we selected 60 image pairs from 7 different sequences.

\subsection{Measuring stereo quality}
\begin{figure}[htb]
    \centering
    \includegraphics[width=\columnwidth]{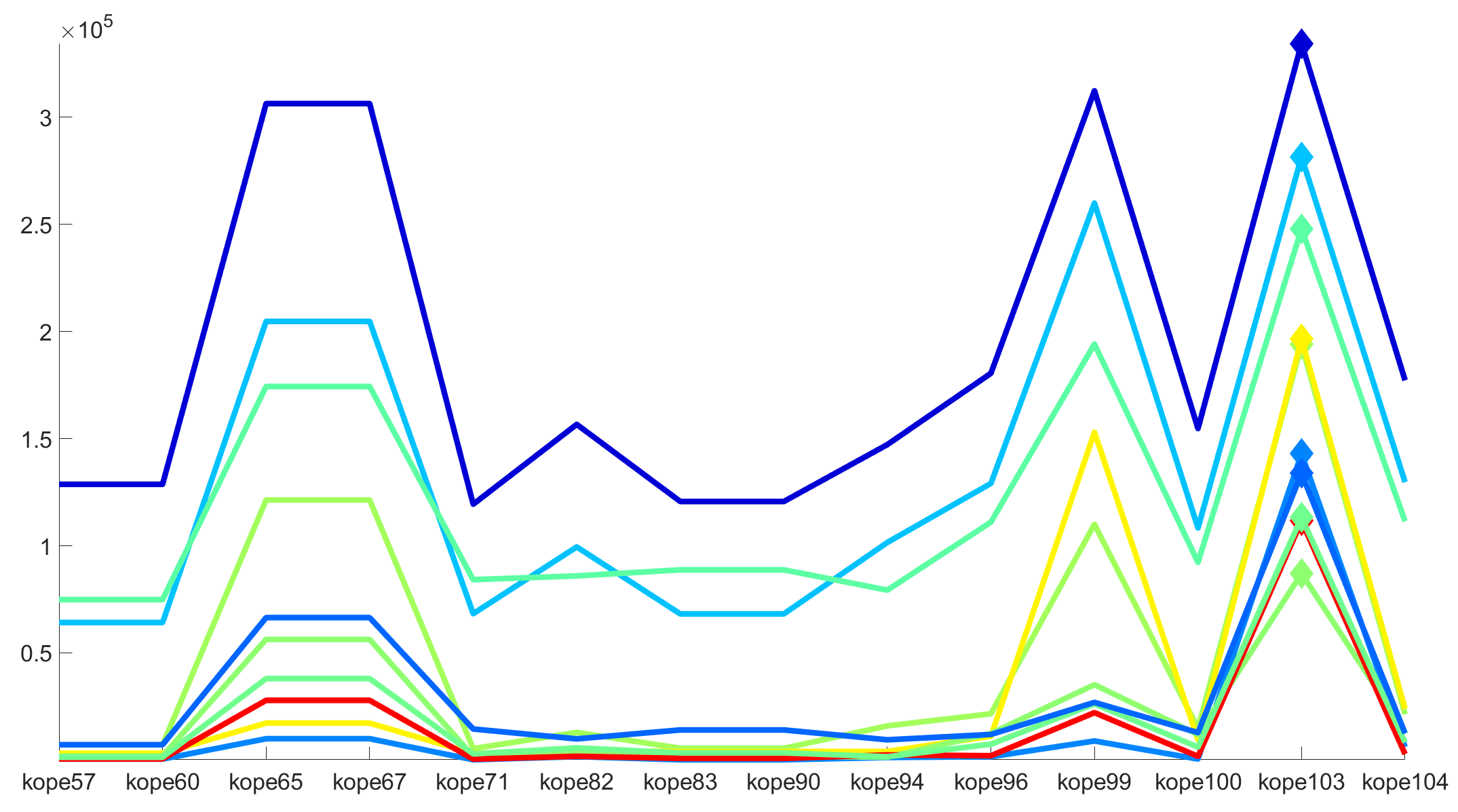}
    \caption{Stereo score for several image pairs (denoted by different colors) from sequence \textit{kope103} using different calibrations (calibration parameter sets). The maximum score for each image pair is marked with a diamond.}
    \label{fig:ex0_graph}
\end{figure}

We first have to show that the number of stereo points can indeed be used as a metric for stereo system calibration quality. A reasonable assumption is that taking the manual calibration that was made right before obtaining image pairs would produce the best results (in terms of quality of disparity images). Consequently, using any other manual calibration would produce disparity images of lower quality. In this experiment, we show that the number of stereo points correlates with our assumptions about calibration quality.
We used 10 image pairs from one sequence (\textit{kope103}) and calculated their stereo score using each available manual calibration. The graph shown in Figure~\ref{fig:ex0_graph} shows the results of the experiment. The $x$ axis shows all the different manual calibrations that were available, whereas the $y$ axis shows the stereo scores calculated from different image pairs (\ie the output of the cost function $f$).

It can be observed that the highest score for all image pairs was achieved when using the calibration parameters calculated right before the mission on which the images were obtained (\textit{kope103}). This shows that the chosen cost function $f$(\ref{eq:f}) reflects the quality of the stereo parameters and is a viable measure of calibration quality. The use of different images from the same mission additionally shows that the metric is independent of the contents of the image pair.

\subsection{Improving manual calibration}
This experiment serves to demonstrate that a decalibrated system can be improved upon by using our approach. To show this we used images from different missions and calculated the stereo score using a random manual calibration that did not correspond to the mission on which the image pair was aquired. The initial scores ($N_i$) were significantly worse than the ground truth ($N_{GT}$), as would be expected. We then used our approach to improve the initial calibration and measured the stereo scores again.

The results are shown in Table~\ref{tab:ex1}. While still worse than the ground truth, all the scores were significantly improved after optimization ($N_o$), which shows potential for automatic recalibration of a stereo system (or at least improving its performance). We also include the metric $r$ which is the ratio between $N_o$ and $N_{GT}$ in the last column. This metric shows how close the score after optimization was to the score of the manual calibration.

\begin{table}[ht]
\resizebox{\columnwidth}{!}{
\begin{tabular}{|ccc|cccc|}
\hline
sequence & image & calibration & $N_i$ & $N_o$ & $N_{GT}$ & $r$ \\ \hline
67 & 9 & 96 & 4895 & 41033 & 90639 & 0.453 \\
67 & 9 & 103 & 78435 & 90646 & 90639 & \textbf{1.000} \\
71 & 5 & 94 & 57160 & 116658 & 111665 & \textbf{1.045} \\
71 & 10 & 94 & 71164 & 168833 & 174877 & 0.965 \\
71 & 10 & 103 & 4498 & 19311 & 174877 & 0.110 \\
103 & 1 & 94 & 15928 & 91393 & 193982 & 0.471 \\
104 & 1 & 57 & 41373 & 109165 & 213418 & 0.512 \\
104 & 4 & 71 & 144585 & 298331 & 415226 & 0.718 \\
104 & 5 & 71 & 76947 & 137803 & 208970 & 0.659 \\
104 & 5 & 99 & 121128 & 195966 & 208970 & 0.938 \\
104 & 6 & 103 & 113290 & 149840 & 237280 & 0.631 \\ \hline
\end{tabular}
}
\label{tab:ex1}
\caption{Stereo scores before and after optimization with different initial calibrations. Boldened values in the last column denote cases where the manual calibration was outperformed.}
\end{table}

\subsection{Ability to generalize across image pairs}
\begin{figure}[htb]
    \centering
    \includegraphics[width=\columnwidth]{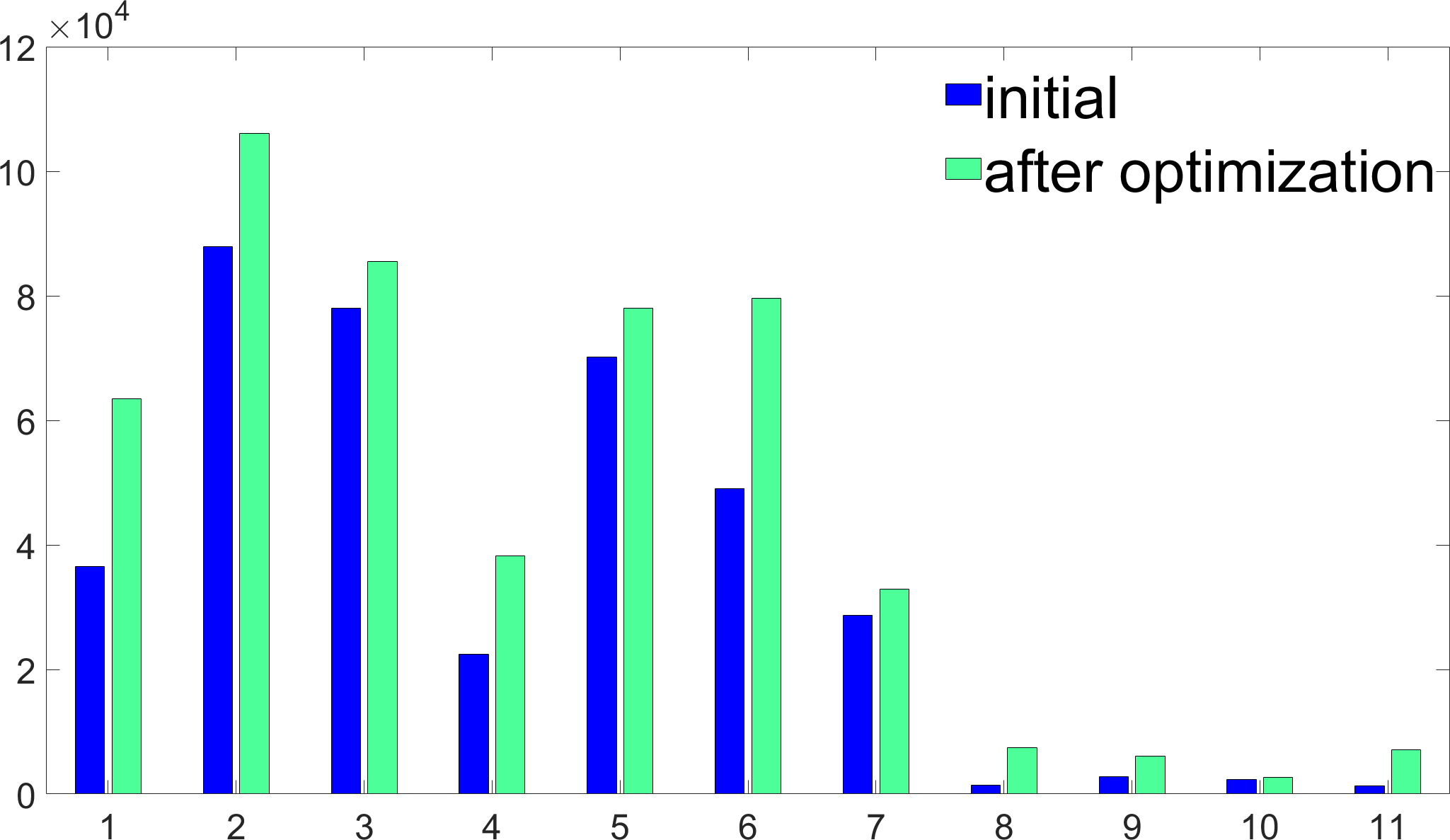}
    \caption{Improvement of the stereo score for different images from sequence \textit{kope104}. The image pair 6 was used for optimization.}
    \label{fig:ex1}
\end{figure}

It could be possible that our optimization only improves the stereo score on the image pair used to perform said optimization. We show that optimization can be performed on one image pair, but the resulting extrinsic parameters also achieve higher stereo scores on other image pairs from the same sequence.
We performed calibration optimization on a randomly chosen image pair from sequence \textit{kope104} with a randomly chosen calibration used as the initial state. We then used the resulting calibration to calculate the stereo score for all remaining images from said sequence and compared the scores to those obtained by using the initial calibration. The results are shown in Figure~\ref{fig:ex1}. We can safely assume all the images from a sequence share the same calibration. The amount of improvement does not seem to be equivalent for all images and all calibrations, but the positive effect is obvious. This also begs the question whether some initial calibrations are more suitable (\ie produce better results) for optimization than others. This prompted us to try and perform optimization with no initial information (c.f.~\ref{exp:ex}). Additionally, this opens up the possibility of sequential (or even online) improvement of the calibration (see further experiments).

\subsection{Fully automatic external calibration}
\label{exp:ex}
Given that different initial calibrations produce results of different qualities, we wanted to make the initial conditions equal for every optimization. Thus we devised an experiment "ex nihilo", where no initial extrinsic information was used, with the exception of the stereo baseline, which was obtained as an average of baseline measurements across all manual calibrations.
The initial parameter set $\Upsilon_i$ is therefore $[0, 0, 0, -356, 0, 0]$, where our baseline is \SI{356}{\milli\meter}.

\begin{figure}[htb]
    \centering
    \includegraphics[width=\columnwidth]{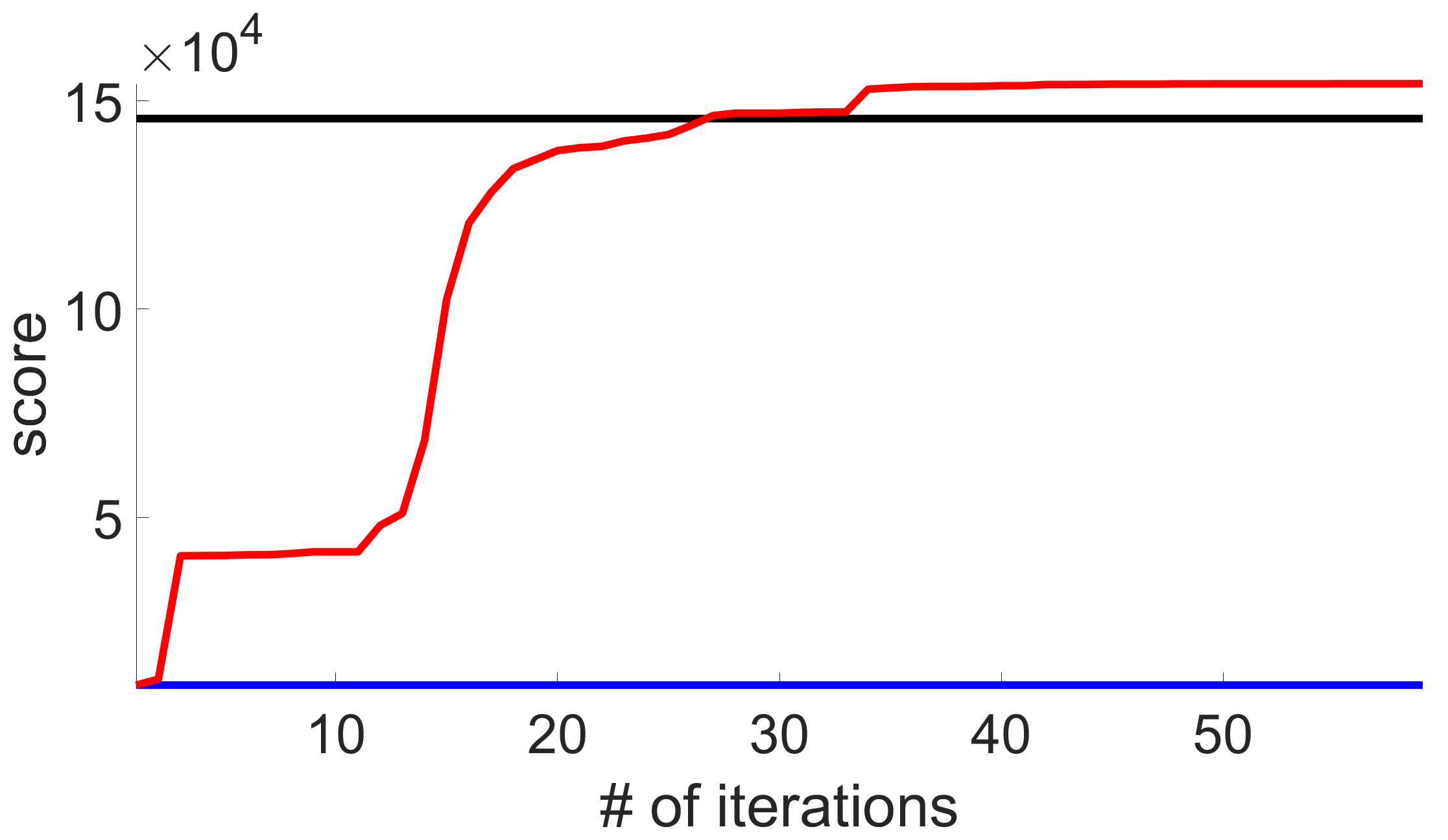}
    \caption{Stereo score during "ex nihilo" calibration. Initial score is shown in blue, while the black line shows the stereo score using the reference calibration. The image pair used was 1 from sequence 104.}
    \label{fig:en_graph}
\end{figure}

\begin{figure}[htb]
    \centering
    \includegraphics[width=\columnwidth]{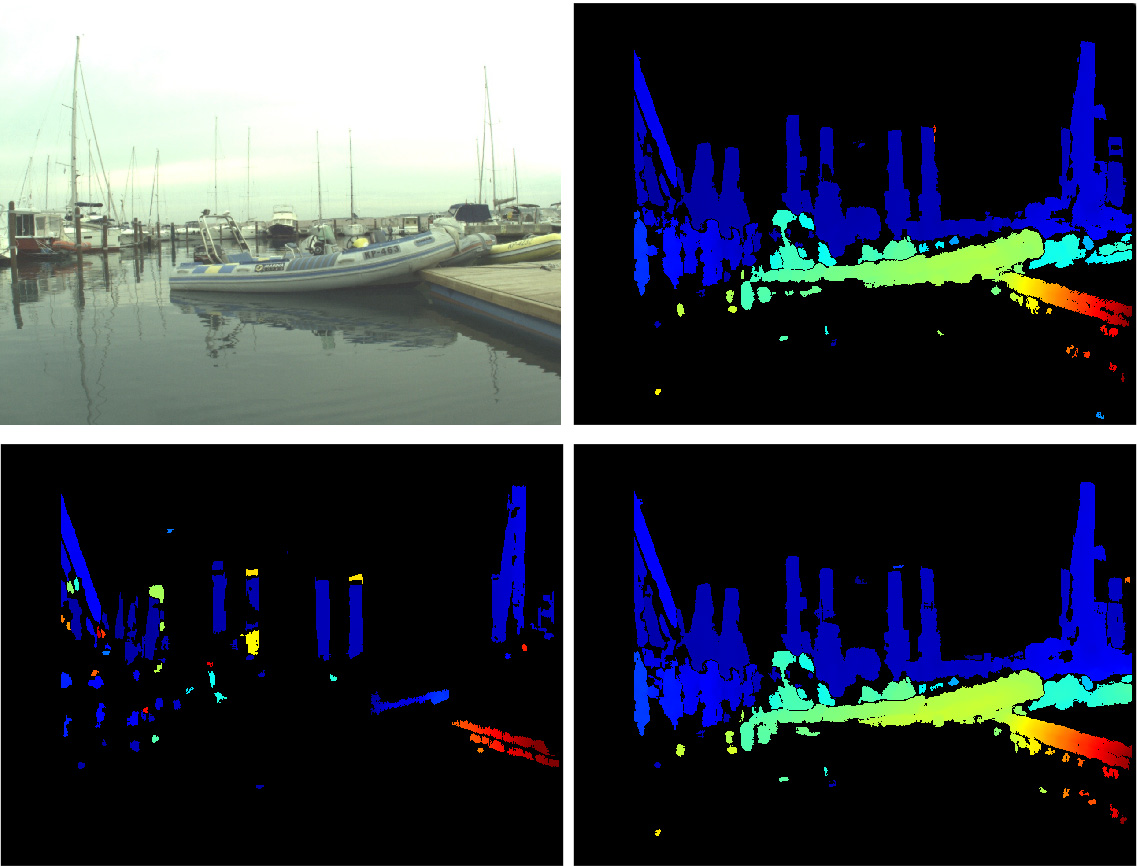}
    \caption{Original image and resulting disparities for image 1 from sequence 104. Upper left: left camera image, upper right: manual calibration, lower left: initial disparity image, lower right: disparity image after optimization.}
    \label{fig:en_im}
\end{figure}

We performed the optimization on three image pairs from seven different sequences. The images were handpicked to be visually diverse with the intent to see how the image content relates to the quality of calibration optimization.

\begin{table}[ht]
\resizebox{\columnwidth}{!}{
\begin{tabular}{|cc|cccc|}
\hline
sequence & image & $N_i$ & $N_o$ & $N_{GT}$ & $r$ \\ \hline
60 & 1 & 164402 & 315856 & 249980 & \textbf{1.264} \\
60 & 2 & 225790 & 304206 & 308937 & 0.985 \\
60 & 3 & 125789 & 255050 & 283294 & 0.900 \\
65 & 1 & 73988 & 119977 & 103851 & \textbf{1.155} \\
65* & 2 & 325230 & 386966 & 378874 & \textbf{1.021} \\
65 & 3 & 345512 & 467605 & 447601 & \textbf{1.045} \\
67 & 1 & 206991 & 236614 & 235365 & \textbf{1.005} \\
67 & 2 & 63970 & 78248 & 77895 & \textbf{1.005} \\
67 & 3 & 173553 & 230362 & 242878 & 0.948 \\
71 & 1 & 9696 & 227088 & 208646 & \textbf{1.088} \\
71* & 2 & 0 & 5241 & 33118 & 0.158 \\
71* & 3 & 489 & 86286 & 82127 & \textbf{1.051} \\
90 & 1 & 113421 & 301904 & 291940 & \textbf{1.034} \\
90 & 2 & 88752 & 338497 & 397074 & 0.852 \\
90* & 3 & 2516 & 177083 & 95813 & \textbf{1.848} \\
103 & 1 & 13654 & 204456 & 193982 & \textbf{1.054} \\
103 & 2 & 90802 & 282952 & 281590 & \textbf{1.005} \\
103 & 3 & 135161 & 339999 & 334506 & \textbf{1.016} \\
104 & 1 & 58423 & 232379 & 213418 & \textbf{1.089} \\
104 & 2 & 104594 & 297483 & 262483 & \textbf{1.133} \\
104 & 3 & 83695 & 362732 & 318165 & \textbf{1.140} \\ \hline
\end{tabular}
}
\caption{Stereo scores after optimization without initial information. Image pairs where we do not expect good results are marked with a star. Boldened values in the last column denote cases where our approach outperformed the manual calibration.}
\label{tab:ex}
\end{table}

The results are shown in Table~\ref{tab:ex}, where the scores are labeled the same as in previous experiments.

Figure~\ref{fig:en_im} depicts the disparities produced by this experiment. It can be observed that the initial parameters produce a quite poor disparity image, whereas the disparity images produced by the manual and optimized calibration are near indistinguishable. The graph in Figure~\ref{fig:en_graph} shows the disparity score (relative to the baseline score, shown in black) with respect to the iteration step for one of the used image pairs.

\begin{figure}[htb]
    \centering
    \includegraphics[width=\columnwidth]{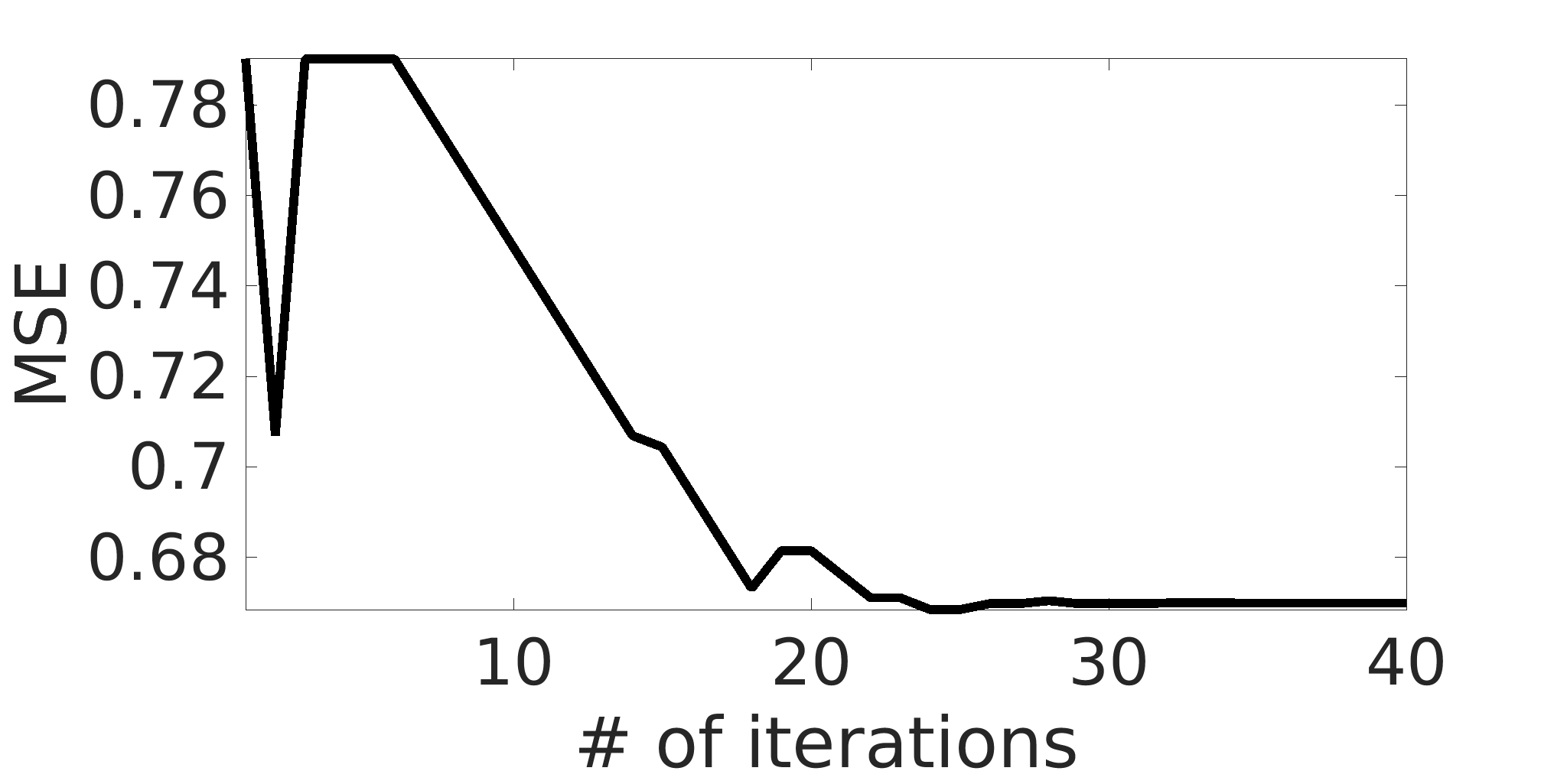}
    \caption{Mean squared error (between ground truth calibration and optimized calibration) during calibration.}
    \label{fig:mse}
\end{figure}
Figure~\ref{fig:mse} additionally shows how the mean square error between the ground truth and the optimization parameters changes during the optimization. As noted in Section~\ref{approach}, the distance to the ground truth generally does not mirror the quality of the optimization because the manual calibration might not be the globally optimal position to begin with.

Results seem promising, with the optimized calibration consistently achieving a result close to or even better than the manual calibration. The few exceptions to this appear because the initial score was too low and optimization had trouble finding a promising direction to move in. It should also be noted that the poorly performing image pairs all seem to be taken when looking towards the open sea, with little to no coast in sight. The absence of large and strongly textured regions seems to have an adverse effect on the optimization. Images used to perform the optimization should thus be chosen with care. Preliminary research also suggests this could be done automaticaly (with either the heading information of the USV, coupled with a map, or using image content analysis, for example employing single view semantic segmentation, such as~\cite{kristan2016fast}).

\subsection{Simulated sequential calibration}
If we further assume that all images in a single USV session have approximately the same calibration (very similar calibration parameter set), we can expect to obtain a higher quality calibration by iteratively performing the optimization on different images from a single run. This has a side effect of testing the real-world situation in which the calibration is performed in parallel to the actual vehicle guidance in on-board computing hardware. In our USV, we have four processing cores, and one of them is mainly not used, therefore we could run the parallel process which would continously optimize the calibration parameter set. The idea is that a parallel process acquires fresh images from the stereo system in predetermined intervals, and performs automatic calibration on them, with the initial values being the output of the previous completed optimization step. Such setup would result in very robust stereo hardware, which could survive even occasional bumps, heavy vibrations or other, unfortunately realistic, mechanical disturbances. However, since we re-use the result of stereo disparity estimation, similar strategy could be employed even without parallel computation, using the ''live'' disparity computation, as long as the scene in front of the vehicle does not rapidly change in the time that is needed for convergence of the optimization method.

\begin{figure}[htb]
    \centering
    \includegraphics[width=\columnwidth]{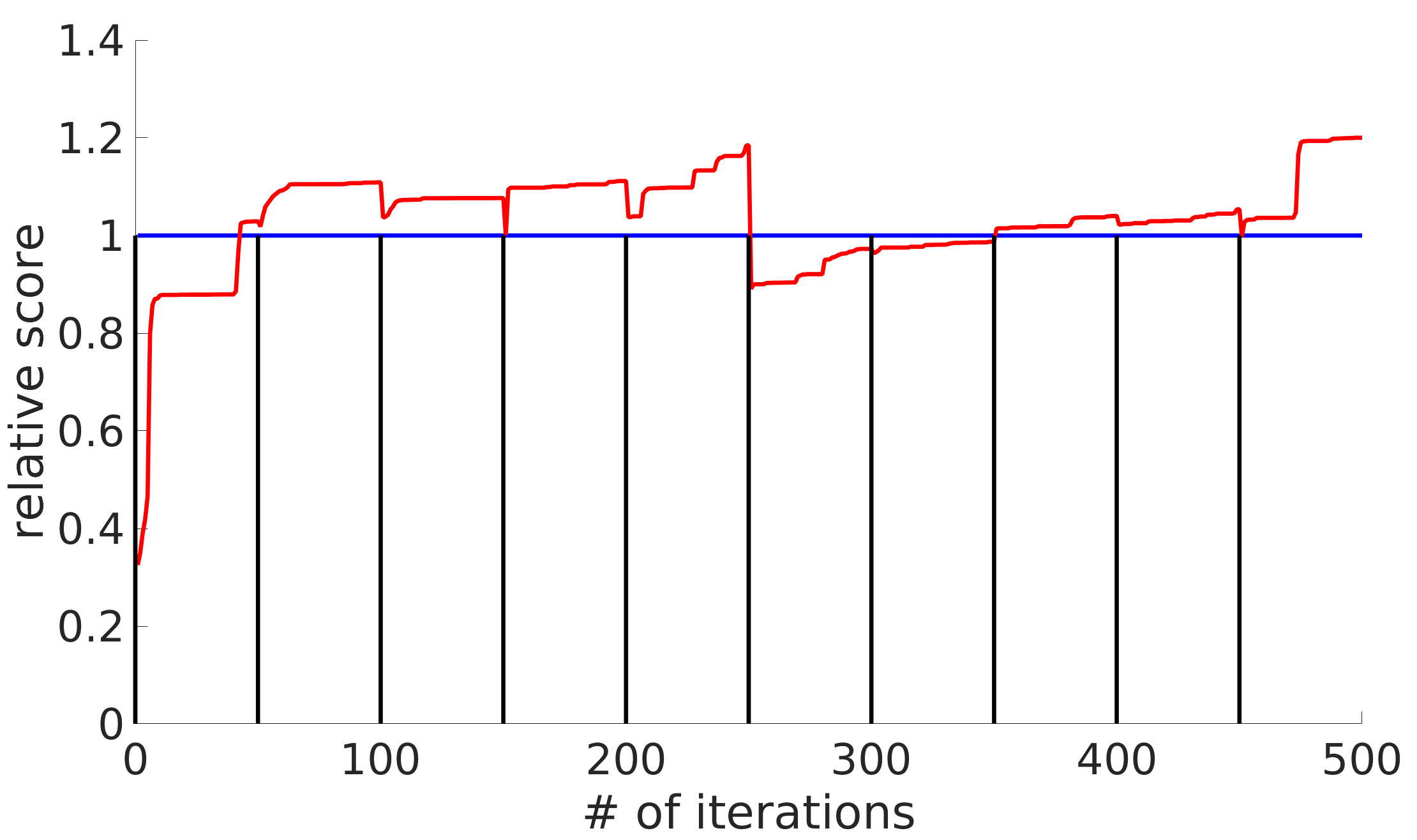}
    \caption{Simulated sequential calibration. A new image pair was introduced into optimization loop at the intervals denoted by the vertical black lines. The stereo score during the optimization is shown in red. The score is scaled to be relative to the reference calibration.}
    \label{fig:cross_graph}
\end{figure}

To simulate such a process, we used a sequence of 10 images. We performed the optimization on all of them in a sequential manner. No initial information was used (same as in Section~\ref{exp:ex}). The order of the images was randomized as to avoid bias in that regard (corresponds to unknown random route of the USV). Figure~\ref{fig:cross_graph} shows the score of the calibration during the optimization for each image. The optimization on each image was set to run for 50 iterations. The image change is denoted by a vertical black line. The score of the optimization is normalized with regard to the score of the manual calibration for the current image. It can be observed that the score keeps improving with time and only sometimes drops when the image is changed. It can be extrapolated that a better choice of images used, a higher number of them, or several passes over all of them could improve the calibration even further. As noted in the previous experiment, image pairs are not all equally suited for optimization. A rough metric by which to choose could simply be the initial stereo score, as even poor initial calibration gives higher scores on images with large, textured objects that usually appear in images that optimization works well with.


\section{Conclusion}
We demonstrated that an indirect measurement of the quality of image rectification for stereo systems can be used for numerical optimization. The number of valid disparity points from a stereo matching algorithm proved to be a good metric over which optimization can be performed. Long history of work with stereo-based USV navigation enabled us to have historical record of calibrations and corresponding image sequences, which enabled us to truthfully simulate real world conditions. The extrinsic parameters of the stereo system were iteratively optimized using the compass search method. We demonstrated that even though decalibration during vehicle operation can occur, our method achieves results that consistently rival the quality of the results produced by a manual calibration. Additionally, the results of our method can at times exceed the baseline score of the manual calibration, thereby improving the quality of the disparity image even further. Additional value of our method can be gained from the fact that we can use it to retroactively calibrate the camera for past USV missions. If the stereo system was not properly calibrated before a mission, the stereo data could be useless. By applying our method, we can calibrate the system for data that was acquired with a poorly calibrated system.

Given the robustness of our method to initial parameters, it could potentially be used instead of a manual calibration. Further work will be in the direction of more analytical approach, and in calibration of the baseline with help of the additional absolute distance sensor (LIDAR).

\section*{Acknowledgement}
This work was in part supported by the Slovenian research agency ARRS programs P2-0214 and P2-0095 and the Slovenian research agency ARRS research project J2-8175.

{\small
\bibliographystyle{ieee}
\bibliography{bibliography}
}

\end{document}